\documentclass[11pt]{article}

\usepackage[utf8]{inputenc}
\usepackage[T1]{fontenc}
\usepackage{amsmath,amssymb}
\usepackage{graphicx}
\usepackage{booktabs}
\usepackage{hyperref}
\usepackage{xcolor}
\usepackage{listings}
\usepackage{algorithm}
\usepackage{algpseudocode}
\usepackage[margin=1in]{geometry}
\usepackage{caption}
\usepackage{subcaption}
\usepackage{enumitem}
\usepackage{natbib}

\definecolor{codebg}{RGB}{245,245,245}
\definecolor{codegreen}{RGB}{40,160,40}
\definecolor{codegray}{RGB}{128,128,128}
\definecolor{codepurple}{RGB}{160,32,240}

\lstdefinestyle{python}{
    backgroundcolor=\color{codebg},
    basicstyle=\ttfamily\small,
    breaklines=true,
    captionpos=b,
    commentstyle=\color{codegreen},
    keywordstyle=\color{codepurple}\bfseries,
    stringstyle=\color{orange},
    numberstyle=\tiny\color{codegray},
    numbers=left,
    numbersep=5pt,
    frame=single,
    framesep=5pt,
    language=Python,
    showstringspaces=false,
    tabsize=4,
    morekeywords={self,True,False,None,as,with,yield,mx,nn,mlxsnn},
}

\title{\textbf{mlx-snn: Spiking Neural Networks on Apple Silicon via MLX}}

\author{
    Jiahao Qin \\
    \texttt{Jiahao.qin19@gmail.com}
}

\date{}

\begin{document}

\maketitle

\begin{abstract}
We introduce \textbf{mlx-snn}, the first spiking neural network (SNN) library built natively on Apple's MLX framework. As SNN research grows rapidly, all major libraries --- snnTorch, Norse, SpikingJelly, Lava --- target PyTorch or custom backends, leaving Apple Silicon users without a native option. mlx-snn provides six neuron models (LIF, IF, Izhikevich, Adaptive LIF, Synaptic, Alpha), four surrogate gradient functions, four spike encoding methods (including an EEG-specific encoder), and a complete backpropagation-through-time training pipeline. The library leverages MLX's unified memory architecture, lazy evaluation, and composable function transforms (\texttt{mx.grad}, \texttt{mx.compile}) to enable efficient SNN research on Apple Silicon hardware. We validate mlx-snn on MNIST digit classification across five hyperparameter configurations and three backends, achieving up to 97.28\% accuracy with 2.0--2.5$\times$ faster training and 3--10$\times$ lower GPU memory than snnTorch on the same M3 Max hardware. mlx-snn is open-source under the MIT license and available on PyPI.
\end{abstract}

\noindent\textbf{Project:} \url{https://github.com/D-ST-Sword/mlx-snn}

\section{Introduction}

Spiking neural networks (SNNs), often called the ``third generation'' of neural networks~\citep{maass1997networks}, process information through discrete spike events rather than continuous activations. This biologically inspired paradigm offers unique advantages: temporal coding of information, event-driven computation with potential energy savings, and natural compatibility with neuromorphic hardware~\citep{davies2018loihi}. Recent advances in surrogate gradient learning~\citep{neftci2019surrogate} have made SNNs trainable with gradient descent, closing the accuracy gap with conventional deep learning on many tasks.

Several software libraries have emerged to support SNN research. snnTorch~\citep{eshraghian2023training} provides a PyTorch-based framework with an intuitive API. Norse~\citep{pehle2021norse} emphasizes functional programming and biological plausibility. SpikingJelly~\citep{fang2023spikingjelly} offers high-performance CUDA kernels. Lava~\citep{lava2021} targets Intel's Loihi neuromorphic processor. However, all these libraries are built on PyTorch~\citep{paszke2019pytorch} or custom backends, leaving users of Apple Silicon hardware without a native SNN framework.

Apple's MLX~\citep{mlx2023} is an array computation framework designed specifically for Apple Silicon. Its unified memory architecture eliminates CPU--GPU data transfers, and its lazy evaluation model enables efficient computation graph optimization. MLX provides composable function transforms --- \texttt{mx.grad} for automatic differentiation and \texttt{mx.compile} for just-in-time compilation --- that are well-suited to SNN workloads requiring custom gradient definitions and unrolled temporal loops.

In this paper, we present \textbf{mlx-snn}, the first SNN library built natively on MLX. Our contributions are:
\begin{enumerate}[nosep]
    \item A comprehensive SNN library with 6 neuron models, 4 surrogate gradient functions, and 4 spike encoding methods, all implemented purely in MLX.
    \item A novel straight-through estimator (STE) pattern for surrogate gradients that works around MLX's current limitations with custom VJP definitions.
    \item An snnTorch-compatible API that enables straightforward migration of existing SNN code to Apple Silicon.
    \item Empirical validation on MNIST classification with comparison to snnTorch.
\end{enumerate}

\section{Related Work}

\paragraph{SNN Libraries.}
Table~\ref{tab:comparison} summarizes the major SNN software frameworks. snnTorch~\citep{eshraghian2023training} is the most widely used, providing neuron models as PyTorch modules with surrogate gradient support. Norse~\citep{pehle2021norse} takes a functional approach inspired by JAX~\citep{jax2018}, separating state from computation. SpikingJelly~\citep{fang2023spikingjelly} focuses on performance, offering fused CUDA kernels for accelerated training. Lava~\citep{lava2021} provides a hardware-software co-design framework targeting Intel's Loihi processor. All of these libraries depend on PyTorch for gradient computation and GPU acceleration via CUDA.

\paragraph{Apple MLX.}
MLX~\citep{mlx2023} is a NumPy-like array framework optimized for Apple Silicon's unified memory architecture. Key features include lazy evaluation (computations are only executed when results are needed), composable function transforms (\texttt{mx.grad}, \texttt{mx.vmap}, \texttt{mx.compile}), and automatic device management. The growing MLX ecosystem includes libraries for language models (mlx-lm), computer vision, and audio processing, but prior to mlx-snn, no SNN library existed for MLX.

\begin{table}[t]
\centering
\caption{Feature comparison of SNN software libraries. $\checkmark$ = supported, $\circ$ = partial, --- = not supported.}
\label{tab:comparison}
\small
\begin{tabular}{lccccc}
\toprule
\textbf{Feature} & \textbf{mlx-snn} & \textbf{snnTorch} & \textbf{Norse} & \textbf{SpikingJelly} & \textbf{Lava} \\
\midrule
Backend & MLX & PyTorch & PyTorch & PyTorch & Custom \\
Neuron models & 6 & 6 & 8 & 10+ & 4 \\
Surrogate gradients & 4 & 5 & 3 & 4 & --- \\
Spike encoding & 4 & 4 & 2 & 3 & --- \\
Apple Silicon native & $\checkmark$ & --- & --- & --- & --- \\
CUDA acceleration & --- & $\checkmark$ & $\checkmark$ & $\checkmark$ & --- \\
Neuromorphic HW & --- & --- & --- & $\circ$ & $\checkmark$ \\
Functional API & $\checkmark$ & $\circ$ & $\checkmark$ & $\circ$ & --- \\
Learnable dynamics & $\checkmark$ & $\checkmark$ & $\checkmark$ & $\checkmark$ & --- \\
Medical encoders & $\checkmark$ & --- & --- & --- & --- \\
\bottomrule
\end{tabular}
\end{table}

\section{Architecture and Design}

\subsection{Design Principles}

mlx-snn follows four core design principles:

\begin{enumerate}[nosep]
    \item \textbf{MLX-native}: All tensor operations use \texttt{mlx.core}. NumPy is used only for data I/O.
    \item \textbf{Explicit state}: Neuron state is passed as a Python dictionary, making the library compatible with MLX's functional transforms and \texttt{mx.compile}.
    \item \textbf{snnTorch-compatible API}: Class names, constructor arguments, and forward-pass signatures mirror snnTorch wherever possible.
    \item \textbf{Research-first}: Every component can be subclassed, overridden, or composed.
\end{enumerate}

\noindent Listing~\ref{lst:api_comparison} illustrates the API similarity between snnTorch and mlx-snn.

\begin{lstlisting}[caption={API comparison: snnTorch (left comments) vs.\ mlx-snn.}, label={lst:api_comparison}]
# --- snnTorch ---                 # --- mlx-snn ---
import snntorch as snn             import mlxsnn
lif = snn.Leaky(beta=0.9)         lif = mlxsnn.Leaky(beta=0.9)
mem = lif.init_leaky()             state = lif.init_state(B, F)
spk, mem = lif(x, mem)            spk, state = lif(x, state)
\end{lstlisting}

\subsection{Neuron Models}

All neuron models inherit from \texttt{SpikingNeuron}, an abstract \texttt{mlx.nn.Module} subclass that provides the \texttt{fire()} and \texttt{reset()} methods. The \texttt{fire()} method applies the surrogate gradient function to the difference between membrane potential and threshold. The \texttt{reset()} method supports three mechanisms: \textit{subtract} (subtract threshold), \textit{zero} (reset to zero), and \textit{none} (no reset, used for output layers).

\paragraph{Leaky Integrate-and-Fire (LIF).}
The standard first-order LIF neuron~\citep{gerstner2002spiking} with discrete-time update:
\begin{equation}
    U[t+1] = \beta \cdot U[t] + X[t+1] - S[t] \cdot V_{\text{thr}}
    \label{eq:lif}
\end{equation}
where $U$ is membrane potential, $X$ is input current, $S \in \{0,1\}$ is the output spike, $\beta \in (0,1)$ is the decay factor, and $V_{\text{thr}}$ is the threshold. The decay factor $\beta$ can optionally be made learnable.

\paragraph{Integrate-and-Fire (IF).}
A non-leaky variant with $\beta = 1$, providing perfect temporal integration.

\paragraph{Izhikevich.}
The two-dimensional Izhikevich model~\citep{izhikevich2003simple}:
\begin{align}
    \frac{dv}{dt} &= 0.04v^2 + 5v + 140 - u + I \label{eq:izh_v} \\
    \frac{du}{dt} &= a(bv - u) \label{eq:izh_u}
\end{align}
with reset rule: if $v \geq 30$ mV, then $v \leftarrow c$ and $u \leftarrow u + d$. Four parameter presets are provided: Regular Spiking (RS), Intrinsically Bursting (IB), Chattering (CH), and Fast Spiking (FS).

\paragraph{Adaptive LIF (ALIF).}
Extends LIF with a spike-frequency adaptation mechanism~\citep{bellec2018long}:
\begin{align}
    U[t+1] &= \beta \cdot U[t] + X[t+1] - S[t] \cdot V_{\text{thr}} \\
    A[t+1] &= \rho \cdot A[t] + S[t] \\
    V_{\text{eff}}[t] &= V_{\text{thr}} + b \cdot A[t]
\end{align}
where $A$ is the adaptation variable, $\rho$ is its decay rate, and $b$ scales the threshold increase.

\paragraph{Synaptic.}
A two-state model with explicit synaptic current filtering:
\begin{align}
    I_{\text{syn}}[t+1] &= \alpha \cdot I_{\text{syn}}[t] + X[t+1] \\
    U[t+1] &= \beta \cdot U[t] + I_{\text{syn}}[t+1] - S[t] \cdot V_{\text{thr}}
\end{align}

\paragraph{Alpha.}
A dual-exponential synapse model cascading two first-order filters for a rise-then-decay post-synaptic current profile:
\begin{align}
    I_{\text{exc}}[t+1] &= \alpha \cdot I_{\text{exc}}[t] + X[t+1] \\
    I_{\text{inh}}[t+1] &= \alpha \cdot I_{\text{inh}}[t] + I_{\text{exc}}[t+1] \\
    U[t+1] &= \beta \cdot U[t] + I_{\text{inh}}[t+1] - S[t] \cdot V_{\text{thr}}
\end{align}

\subsection{Surrogate Gradients}
\label{sec:surrogates}

Spike generation applies the Heaviside step function $\Theta(U - V_{\text{thr}})$, which has zero gradient almost everywhere. Surrogate gradient methods~\citep{neftci2019surrogate, zenke2021remarkable} replace the backward pass with a smooth approximation $\tilde{\sigma}(x)$, enabling gradient-based optimization.

\paragraph{STE pattern.}
MLX's current \texttt{mx.custom\_function} API introduces shape inconsistencies when composed with matrix multiplication backward passes. We adopt a straight-through estimator (STE) pattern using \texttt{mx.stop\_gradient}:
\begin{equation}
    \text{output} = \underbrace{\texttt{stop\_grad}\big(\Theta(x) - \tilde{\sigma}(x)\big)}_{\text{zero gradient}} + \tilde{\sigma}(x)
    \label{eq:ste}
\end{equation}
In the forward pass, this evaluates to $\Theta(x)$ (exact Heaviside). In the backward pass, gradients flow through $\tilde{\sigma}(x)$ only. This pattern is numerically equivalent to a proper custom VJP but avoids the shape bug.

\paragraph{Supported functions.}
mlx-snn provides three built-in surrogate functions plus a custom surrogate factory:
\begin{itemize}[nosep]
    \item \textbf{Fast sigmoid}: $\tilde{\sigma}(x) = \frac{kx}{2(1+k|x|)} + \frac{1}{2}$, gradient $\frac{k}{2(1+k|x|)^2}$
    \item \textbf{Arctan}: $\tilde{\sigma}(x) = \frac{1}{\pi}\arctan(\alpha x) + \frac{1}{2}$
    \item \textbf{Straight-through}: $\tilde{\sigma}(x) = \text{clip}(sx + 0.5, 0, 1)$
    \item \textbf{Custom}: User-defined smooth function wrapped in the STE pattern
\end{itemize}

\subsection{Spike Encoding}

mlx-snn provides four spike encoding methods:

\begin{itemize}[nosep]
    \item \textbf{Rate coding}: Poisson spike generation where each input value is interpreted as a firing probability per timestep.
    \item \textbf{Latency coding}: Time-to-first-spike encoding where larger values produce earlier spikes. Supports linear and exponential mappings.
    \item \textbf{Delta modulation}: Change-based encoding that generates spikes when $|x[t] - x[t-1]| > \theta$, suitable for temporal signals.
    \item \textbf{EEG encoder}: A medical-signal-specific encoder supporting rate, delta, and threshold-crossing methods for multi-channel EEG data.
\end{itemize}

All encoders produce time-first tensors of shape $[T, B, \ldots]$, following neuroscience conventions.

\subsection{Training Pipeline}

mlx-snn integrates with MLX's standard training patterns. The \texttt{bptt\_forward} utility unrolls a model over $T$ timesteps, collecting output spikes and membrane potentials. Three loss functions are provided:

\begin{itemize}[nosep]
    \item \textbf{Rate coding loss}: Cross-entropy on spike counts summed over time.
    \item \textbf{Membrane loss}: Cross-entropy on the final-timestep membrane potential (used in our experiments).
    \item \textbf{MSE count loss}: Mean squared error between spike counts and targets.
\end{itemize}

\noindent Listing~\ref{lst:training} shows a complete training loop.

\begin{lstlisting}[caption={Simplified mlx-snn training loop for MNIST classification. \texttt{SpikingMLP} wraps two \texttt{nn.Linear} layers and two \texttt{mlxsnn.Leaky} neurons in a 784--128--10 feedforward architecture. Hyperparameters shown are for illustration; see Table~\ref{tab:mnist} for tuned configurations.}, label={lst:training}]
import mlx.core as mx
import mlx.nn as nn
import mlx.optimizers as optim
import mlxsnn

class SpikingMLP(nn.Module):
    def __init__(self, num_steps=25, beta=0.9):
        super().__init__()
        self.fc1 = nn.Linear(784, 128)
        self.lif1 = mlxsnn.Leaky(beta=beta)
        self.fc2 = nn.Linear(128, 10)
        self.lif2 = mlxsnn.Leaky(beta=beta,
                     reset_mechanism="none")
        self.num_steps = num_steps

    def __call__(self, spikes_in):
        s1 = self.lif1.init_state(
            spikes_in.shape[1], 128)
        s2 = self.lif2.init_state(
            spikes_in.shape[1], 10)
        for t in range(self.num_steps):
            x = self.fc1(spikes_in[t])
            spk, s1 = self.lif1(x, s1)
            x = self.fc2(spk)
            _, s2 = self.lif2(x, s2)
        return s2["mem"]  # final membrane

model = SpikingMLP(num_steps=25, beta=0.9)
optimizer = optim.Adam(learning_rate=1e-3)

def loss_fn(model, spikes_in, targets):
    mem_out = model(spikes_in)  # [batch, 10]
    return mx.mean(
        nn.losses.cross_entropy(mem_out, targets))

loss_and_grad = nn.value_and_grad(model, loss_fn)

for x_batch, y_batch in get_batches(x_train, y_train):
    spikes = mlxsnn.rate_encode(x_batch, num_steps=25)
    loss, grads = loss_and_grad(model, spikes, y_batch)
    optimizer.update(model, grads)
    mx.eval(model.parameters(), optimizer.state)
\end{lstlisting}

\section{MLX-Specific Design Considerations}

\paragraph{Unified memory.}
Apple Silicon's unified memory architecture means that CPU and GPU share the same physical memory. Unlike PyTorch, where \texttt{tensor.to(device)} copies are needed, MLX tensors are accessible to both CPU and GPU without explicit transfers. For SNN workloads that mix data preprocessing (CPU) with neuron simulation (GPU), this eliminates a common bottleneck.

\paragraph{Lazy evaluation.}
MLX uses lazy evaluation: operations build a computation graph that is only executed when \texttt{mx.eval()} is called. This is advantageous for SNNs because the temporal unrolling loop can build the full computation graph for $T$ timesteps before executing, allowing the MLX runtime to optimize memory allocation and kernel scheduling.

\paragraph{Composable transforms.}
MLX's \texttt{mx.grad} transform computes gradients of arbitrary functions, enabling clean BPTT without framework-specific ``retain graph'' semantics. \texttt{mx.compile} can JIT-compile pure functions for additional performance. The explicit state-dict design of mlx-snn neurons ensures compatibility with these transforms --- unlike hidden mutable state, dictionaries compose naturally with functional transformations.

\paragraph{Immutable arrays.}
MLX arrays do not support in-place operations. All neuron updates use the pattern \texttt{mem = beta * mem + x} rather than \texttt{mem += x}. While this requires minor adjustments for users coming from PyTorch, it aligns well with SNN dynamics where the update equations are naturally expressed as functional assignments.

\section{Experiments}

All experiments are run on an Apple M3 Max MacBook Pro (36\,GB unified memory, macOS Sequoia). Software versions: mlx-snn 0.2.1, MLX 0.29.3, snnTorch 0.9.4, PyTorch 2.8.0, Python 3.9.18. We benchmark snnTorch on both the PyTorch MPS backend (Apple Silicon GPU) and CPU to provide a comprehensive comparison.

\subsection{MNIST Classification}

\paragraph{Setup.}
We train a two-layer feedforward SNN on MNIST~\citep{lecun1998gradient}: $784 \rightarrow h$ (LIF) $\rightarrow 10$ (LIF, no reset). Input images are rate-encoded into Poisson spike trains over $T=25$ timesteps. The output layer accumulates membrane potential without reset, and classification uses the argmax of the final membrane state. We sweep five hyperparameter configurations (Table~\ref{tab:mnist}) covering decay factors $\beta \in \{0.85, 0.9, 0.95\}$, hidden sizes $h \in \{128, 256\}$, learning rates $\{10^{-3}, 2{\times}10^{-3}\}$, and batch sizes $\{128, 256\}$, and evaluate on three backends: mlx-snn (MLX GPU), snnTorch (PyTorch MPS GPU), and snnTorch (PyTorch CPU). All use the Adam optimizer and fast sigmoid surrogate gradient.

\begin{figure}[t]
\centering
\includegraphics[width=\textwidth]{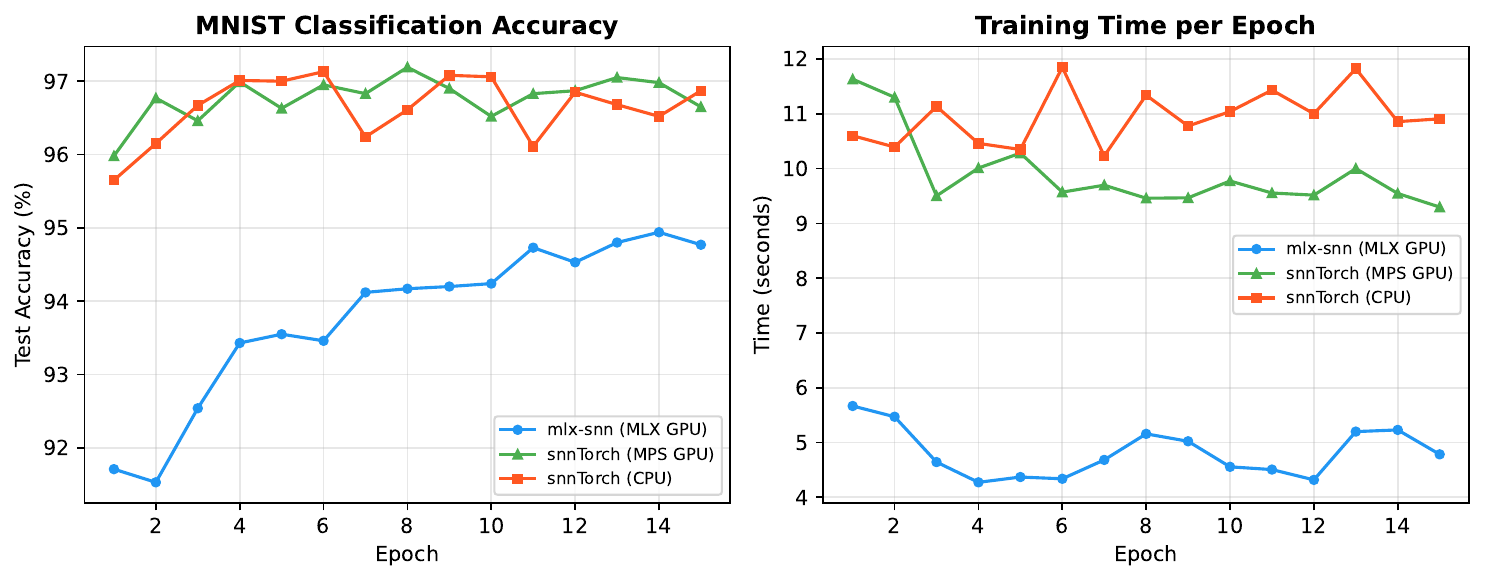}
\caption{MNIST training curves for configuration C5 ($\beta{=}0.95$, $h{=}128$). Left: test accuracy vs.\ epoch. Right: training time per epoch.}
\label{fig:training_curves}
\end{figure}

\paragraph{Results.}
Table~\ref{tab:mnist} reports best test accuracy and average training time per epoch across all 15 runs. mlx-snn is consistently 2.0--2.5$\times$ faster per epoch than snnTorch on both MPS and CPU backends, while using 3.3--10.5$\times$ less GPU memory. The best mlx-snn configuration (C1: $\beta{=}0.85$, $h{=}256$, lr$=10^{-3}$) achieves 97.28\% accuracy, within 0.7 points of snnTorch's best (98.03\%). The accuracy gap is consistent across all configurations (${\sim}$1 percentage point) and is attributable to the STE-based surrogate gradient pattern (Eq.~\ref{eq:ste}), which routes gradients through a smooth approximation rather than a native custom backward pass. Notably, snnTorch achieves similar accuracy on both MPS and CPU, confirming that the gap stems from the surrogate gradient formulation rather than the compute backend. We expect this gap to narrow as MLX's \texttt{mx.custom\_function} matures. Figure~\ref{fig:training_curves} shows the training curves for configuration C5.

\begin{table}[t]
\centering
\caption{MNIST classification: 5 hyperparameter configurations $\times$ 3 backends. All use the same $784$--$h$--$10$ feedforward SNN with $T{=}25$ timesteps, fast sigmoid surrogate, and batch size 128 unless noted. C1--C4 train for 25 epochs; C5 trains for 15 epochs. Acc.\ = best test accuracy over all epochs; Time = average seconds per epoch; Mem.\ = peak GPU allocator memory (Metal for mlx-snn, MPS for snnTorch).}
\label{tab:mnist}
\small
\begin{tabular}{llcccccc}
\toprule
& & \multicolumn{2}{c}{\textbf{mlx-snn (MLX)}} & \multicolumn{2}{c}{\textbf{snnTorch (MPS)}} & \multicolumn{2}{c}{\textbf{snnTorch (CPU)}} \\
\cmidrule(lr){3-4} \cmidrule(lr){5-6} \cmidrule(lr){7-8}
& \textbf{Hyperparameters} & Acc.\,(\%) & Time\,(s) & Acc.\,(\%) & Time\,(s) & Acc.\,(\%) & Time\,(s) \\
\midrule
C1 & $\beta{=}0.85,\; h{=}256,\; \text{lr}{=}10^{-3}$ & 97.28 & \textbf{4.0} & 98.00 & 8.8 & \textbf{98.01} & 12.8 \\
C2 & $\beta{=}0.9,\; h{=}256,\; \text{lr}{=}10^{-3}$ & 97.02 & \textbf{4.3} & \textbf{98.03} & 8.9 & 97.97 & 13.5 \\
C3 & $\beta{=}0.9,\; h{=}256,\; \text{lr}{=}10^{-3},\; B{=}256$ & 96.91 & \textbf{2.4} & 98.03 & 4.8 & \textbf{98.17} & 16.7 \\
C4 & $\beta{=}0.9,\; h{=}128,\; \text{lr}{=}10^{-3}$ & 96.90 & \textbf{4.3} & \textbf{97.84} & 9.0 & 97.74 & 10.9 \\
C5 & $\beta{=}0.95,\; h{=}128,\; \text{lr}{=}2{\times}10^{-3},\; 15\text{ep}$ & 94.98 & \textbf{4.4} & \textbf{97.09} & 9.0 & 97.00 & 11.1 \\
\midrule
\multicolumn{2}{l}{\textit{Peak GPU memory}} & \multicolumn{2}{c}{61--138\,MB} & \multicolumn{2}{c}{241--1453\,MB} & \multicolumn{2}{c}{---} \\
\bottomrule
\end{tabular}
\end{table}

\subsection{Neuron Dynamics Validation}

\begin{figure}[t]
\centering
\includegraphics[width=\textwidth]{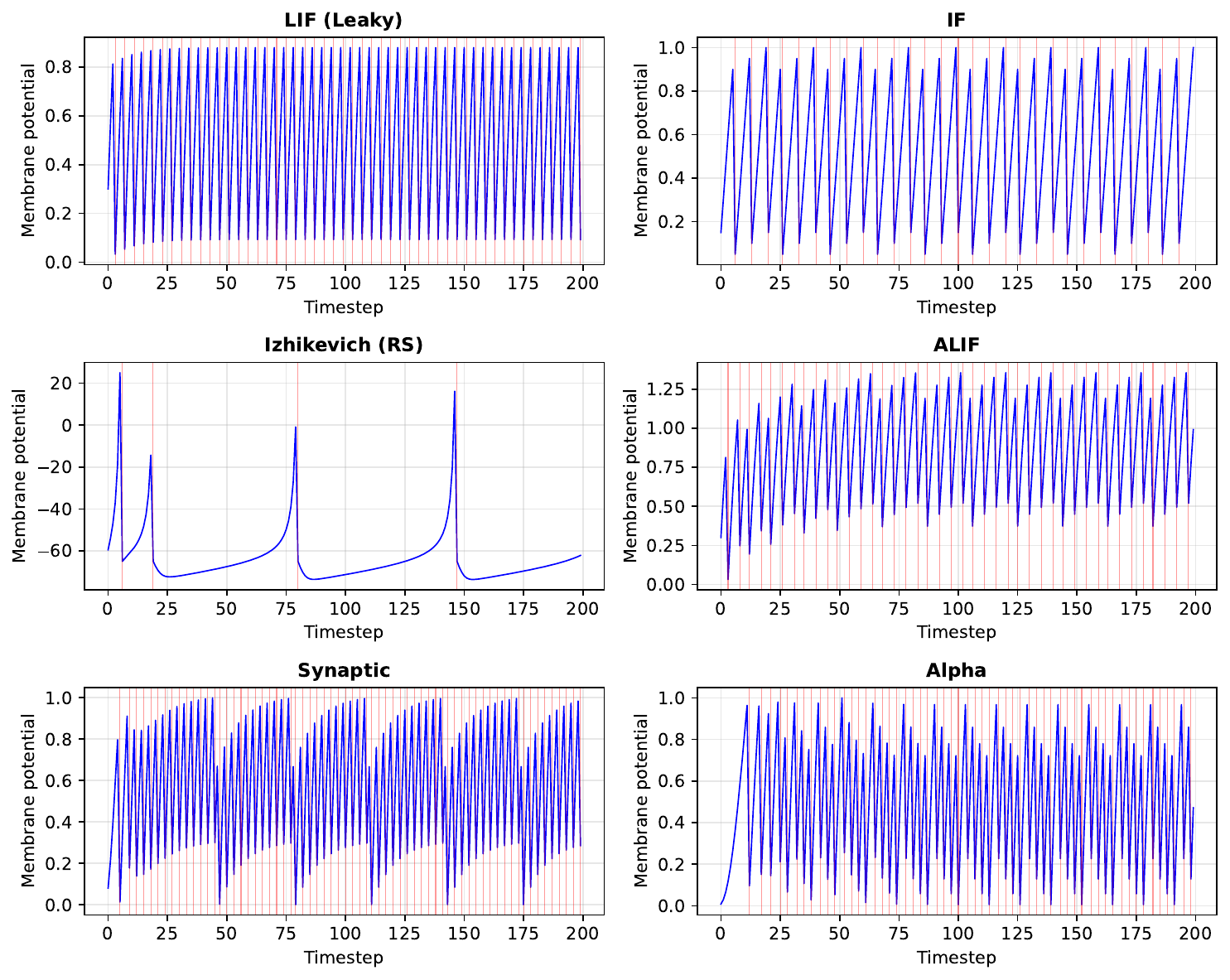}
\caption{Membrane potential traces for all six neuron models with constant input current over 200 timesteps. Red vertical lines indicate spike events.}
\label{fig:neuron_dynamics}
\end{figure}

Figure~\ref{fig:neuron_dynamics} shows membrane potential traces for all six neuron models under constant input current. Each model exhibits its characteristic dynamics: LIF shows exponential rise to threshold and reset; IF accumulates linearly; Izhikevich displays the distinctive quadratic rise and recovery variable interaction; ALIF shows increasing inter-spike intervals due to threshold adaptation; Synaptic shows smoothed dynamics from the synaptic current filter; and Alpha exhibits the rise-then-decay current profile.

\subsection{Surrogate Gradient Comparison}

\begin{figure}[t]
\centering
\includegraphics[width=\textwidth]{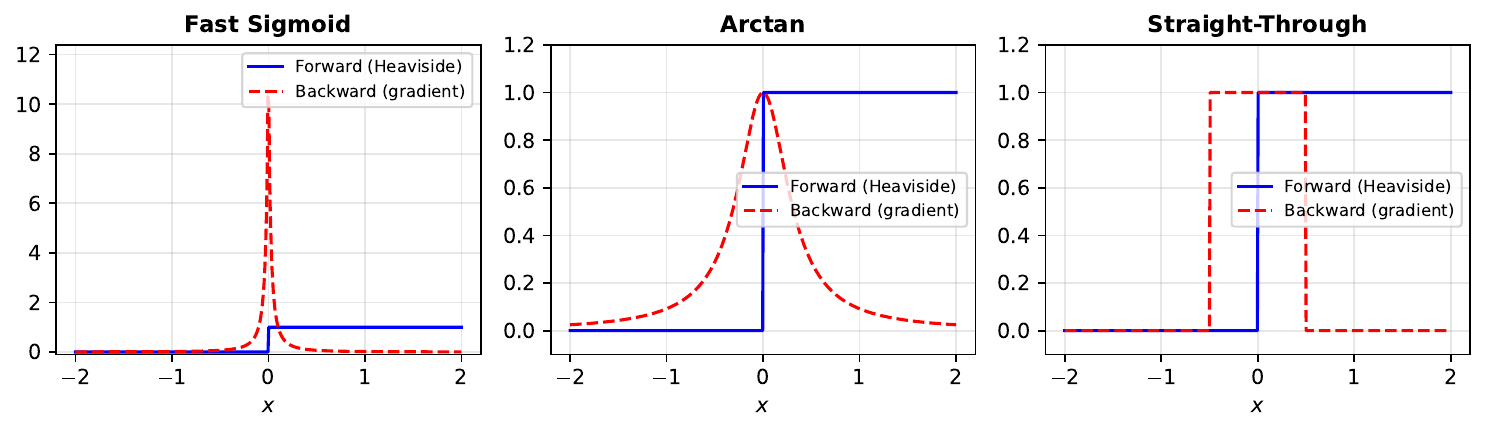}
\caption{Surrogate gradient functions. Blue: forward pass (Heaviside step function). Red dashed: backward pass (smooth gradient used for BPTT).}
\label{fig:surrogates}
\end{figure}

Table~\ref{tab:surrogates} compares three surrogate gradient functions on the MNIST task using the baseline configuration from C5 ($\beta{=}0.95$, $h{=}128$, lr$=2{\times}10^{-3}$, 10 epochs). Fast sigmoid and arctan achieve comparable accuracy (93.65\% and 92.44\%, respectively), consistent with findings by \citet{zenke2021remarkable} that surrogate gradient learning is robust to the choice of surrogate function. The straight-through estimator (46.28\%) underperforms due to its narrow gradient window at the default scale. Figure~\ref{fig:surrogates} visualizes the forward and backward passes for each surrogate.

\begin{table}[t]
\centering
\caption{Surrogate gradient comparison on MNIST using baseline configuration ($\beta{=}0.95$, $h{=}128$, lr$=2{\times}10^{-3}$, 10 epochs). Note: this matches C5 in Table~\ref{tab:mnist}, not the best configuration C1.}
\label{tab:surrogates}
\begin{tabular}{lcc}
\toprule
\textbf{Surrogate Function} & \textbf{Test Accuracy (\%)} & \textbf{Training Time (s)} \\
\midrule
Fast Sigmoid ($k=25$) & \textbf{93.65} & 44.7 \\
Arctan ($\alpha=2$) & 92.44 & 43.8 \\
Straight-Through ($s=1$) & 46.28 & 46.3 \\
\bottomrule
\end{tabular}
\end{table}

\section{Discussion and Future Work}

\paragraph{Current limitations.}
mlx-snn is in active development (v0.2.1). Key limitations include: (1) \texttt{mx.compile} is not yet applied to the training loop due to Python control flow in temporal unrolling; (2) neuromorphic dataset loaders (N-MNIST, DVS-Gesture, SHD) are not yet implemented; (3) the library has been validated only on MNIST --- larger-scale benchmarks are needed.

\paragraph{MLX framework maturity.}
We encountered a shape inconsistency in MLX's \texttt{mx.custom\_function} VJP mechanism when composing with matrix multiplication backward passes. Our STE workaround (Eq.~\ref{eq:ste}) is functionally equivalent but represents a temporary solution. We expect this issue to be resolved as MLX matures, at which point native custom VJPs will provide cleaner surrogate gradient definitions.

\paragraph{Roadmap.}
Planned features for future versions include:
\begin{itemize}[nosep]
    \item \textbf{v0.3.0}: Liquid State Machine (LSM) with configurable reservoir topology, excitatory/inhibitory balance, and EEG classification examples.
    \item \textbf{v0.4.0}: \texttt{mx.compile}-optimized forward passes, neuromorphic dataset loaders, visualization utilities, and comprehensive benchmarks.
    \item \textbf{v1.0.0}: Full API documentation, numerical validation against snnTorch reference outputs, and PyPI stable release.
\end{itemize}

\paragraph{Broader impact.}
By bringing SNN research to Apple Silicon, mlx-snn enables researchers using MacBook Pro or Mac Studio hardware to run SNN experiments without requiring NVIDIA GPUs or cloud infrastructure. The unified memory architecture is particularly advantageous for SNN workloads that require frequent state updates across timesteps. We hope mlx-snn lowers the barrier to entry for SNN research in the Apple ecosystem.

\section{Conclusion}

We have presented mlx-snn, the first spiking neural network library built natively on Apple's MLX framework. The library provides six neuron models, four surrogate gradient functions, and four spike encoding methods with an API designed for compatibility with snnTorch. Our experiments validate the library's correctness on MNIST classification across five configurations and three backends, achieving up to 97.28\% accuracy with 2.0--2.5$\times$ faster training and 3--10$\times$ lower GPU memory compared to snnTorch on the same M3 Max hardware. mlx-snn is open-source under the MIT license and available on PyPI (\texttt{pip install mlx-snn}). To cite this work:

\begin{verbatim}
@article{qin2026mlxsnn,
  title   = {mlx-snn: Spiking Neural Networks
             on Apple Silicon via {MLX}},
  author  = {Qin, Jiahao},
  journal = {arXiv preprint arXiv:XXXX.XXXXX},
  year    = {2026}
}
\end{verbatim}

\bibliographystyle{plainnat}

\end{document}